\documentclass[11pt]{article}
\usepackage{latexsym,amssymb,amsmath} 
\usepackage{path}
\usepackage{url}
\usepackage{verbatim}

\usepackage{natbib}
\usepackage{amsfonts}
\usepackage{amsfonts}
\usepackage{amsfonts}
\usepackage{array}
\usepackage{mathrsfs}
\usepackage{amsfonts}
\usepackage{amsmath}
\usepackage{amssymb}
\usepackage{amsthm}
\usepackage{bm}
\usepackage{appendix}

\usepackage{graphicx} 
\usepackage{subfigure}

\graphicspath{{figures/}}

\usepackage[lined,boxed,ruled, commentsnumbered]{algorithm2e}


\numberwithin{equation}{section}
\numberwithin{table}{section}
\numberwithin{figure}{section}

\newcommand{\junk}[1]{{}}

\newlength{\fwtwo} 

\title{Fedlearn-Algo: A flexible open-source privacy-preserving machine learning platform}
\author{
Bo Liu, Chaowei Tan, Jiazhou Wang, Tao Zeng, Huasong Shan, \\
Houpu Yao, Heng Huang, Peng Dai, Liefeng Bo, Yanqing Chen \\ \\
JD Finance America Corporation\\ Mountain View, CA, USA \\
\{bo.liu2, chaowei.tan, jiazhou.wang3, tao.zeng, huasong.shan, houpu.yao, \\heng.huang, peng.dai, liefeng.bo, yanqing.chen\}@jd.com \\
\date{}
}

\begin{document}

\maketitle
\begin{abstract}
    In this paper, we present Fedlearn-Algo, an open-source privacy preserving machine learning platform. We use this platform to demonstrate our research and development results on privacy preserving machine learning algorithms. As the first batch of novel FL algorithm examples, we release vertical federated kernel binary classification model and vertical federated random forest model. They have been tested to be more efficient than existing vertical federated learning models in our practice. Besides the novel FL algorithm examples, we also release a machine communication module. The uniform data transfer interface supports transferring widely used data formats between machines. We will maintain this platform by adding more functional modules and algorithm examples. The code is available at \url{https://github.com/fedlearnAI/fedlearn-algo}.
\end{abstract}
\section{Introduction}
Powerful AI model is built upon learning from sufficient training data. However, in many cases, the data owned by one data collector is insufficient to make an AI model well trained, leading to low overall model performance or model bias. One solution is increasing the training data scale by utilizing the data from different parties. This is a common solution to many use cases where the data from multiple sources are complementary. For example,  customer can have purchasing and browsing history on multiple E-commerce platforms. Product recommendation models trained on all these data can definitely outperform models trained by each platform on its own data~\citep{hu2019fdml}. In medical image analysis, data insufficiency is
a common limitation for high performance AI model development. Emerging efforts are seen to collaboratively use the data from multiple health care institutions for joint model training and the benefits have been demonstrated on various tasks in the literature~\citep{brisimi2018federated,rieke2020future,xu2021federated}.

To build a feasible machine learning solution to cross-device or cross-platform data use, a desirable algorithm has to address the following challenges
\begin{itemize}
\item \textbf{Data privacy protection}. Arbitrary data sharing tends to leak sensitive information like consumer privacy, leading to unpredictable future risk and hurting the customers' trust towards the data controller. Data privacy protection is progressively enforced by government legislation. GDPR requires a data protection impact assessment (DPIA) for any data use\footnote{\url{https://gdpr.eu/data-protection-impact-assessment-template/}}. The assessment includes solving privacy risk. The data use for AI model learning purpose also subjects to this regulation.
\item \textbf{Communication cost}. The time cost of a multi-machine algorithm mainly comes from local computation and machine communication. Since currently there have been multiple ways to speedup the computation on single machine (e.g. parallel computing, well-studied efficient single machine model training algorithms), the major bottleneck is the machine communication cost. Communication time cost depends on a number of highly uncontrollable factors such as network workload, network topology and the overall workload of each machine, etc. Popular large models have millions or even billions of parameters, transferring float point numbers at such scale on public network environment takes a long time. Considering the iterative nature of multi-machine algorithms, the overall communication cost can be prohibitive.

\item \textbf{Algorithm performance}. The complicated multi-machine data properties and machine collaboration mechanism produce many new algorithm research issues. Several problems have aroused extensive research attention, such as data statistical heterogeneity~\citep{nishio2019client} and data imbalance~\citep{duan2020self}, etc. Those issues are closely related to the model performance. To fully exploit the value of data in model learning, they have to be considered in algorithm design, deserving further research efforts.

\end{itemize}

Federated Learning (FL) is among the emerging efforts that target at the above challenges. It was initially proposed by Google as an solution to using data from multiple mobile devices for next word prediction model learning~\citep{mcmahan2017communication}. The idea soon gains extensive attention from both industry and academia due to its significant practical value and the numerous research issues waiting to be solved. According to the data partition differences, most of existing FL algorithms can be mainly categorized into horizontal FL algorithms and vertical FL algorithms~\citep{yang2019federated}. Horizontal FL refers to the setting that samples on the involved machines share the same feature space while the machines have different sample ID space. Vertical FL refers to the setting that all machines share the same sample ID space and each machine has a unique feature space.

Deploying a multi-machine algorithm is known to be more challenging than single machine algorithm as far as algorithm design and analysis, implementation, debugging and testing are concerned. In this work, we present Fedlearn-Algo, an open-source FL algorithm platform. We release this tool as a platform to demonstrate our current and future privacy-preserving machine learning algorithm research results. Meanwhile, we believe the extensible and flexible overall framework design make it helpful to FL research community by which a multi-machine algorithm can be
easily developed. Specifically, Fedlearn-Algo is characterized by the following highlights.
\begin{itemize}
    \item \textbf{Novel vertical FL algorithms.} Most existing FL open-source softwares (e.g. FedML\footnote{\url{https://fedml.ai/}}, Flower\footnote{\url{https://flower.dev/}}, TensorFlow Federated\footnote{\url{https://www.tensorflow.org/federated}}, etc.) and algorithm research efforts are mainly dedicated in horizontal FL algorithm development. Vertically partitioned data is seen in many to Business (toB) and Government (toG) applications. Despite the existing vertical FL models such as SecureBoost~\citep{cheng2021secureboost} and homomorphic encryption based logistic regression model~\citep{hardy2017private}, their efficiency are found to be unsatisfactory in our real-world FL deployment practice. This motivates us design  novel vertical FL algorithms including vertical federated kernel method and vertical federated random forest model. We release prototype of these algorithms. In the future we will release more vertical FL algorithm design results.

    \item \textbf{Easy-to-use machine communication module.} Besides the released vertical FL algorithms, we believe the communication module serving all released algorithms is also friendly to contributors or researchers for their multi-machine algorithm implementation. The information format, parameter number and parameter size transferred between machines differ in FL algorithms. We design a uniform message data structure. It supports the widely used data formats (e.g. int, string, float, vector, matrix, etc.) and arbitrary number of parameters to be transferred within one message. Developers can use it conveniently in their own algorithm implementation for transferred message definition. An uniform message transfer interface is provided to transfer the message. 


\end{itemize}

\section{Platform Overview}
\begin{figure}
\centering
\includegraphics[width=1\textwidth]{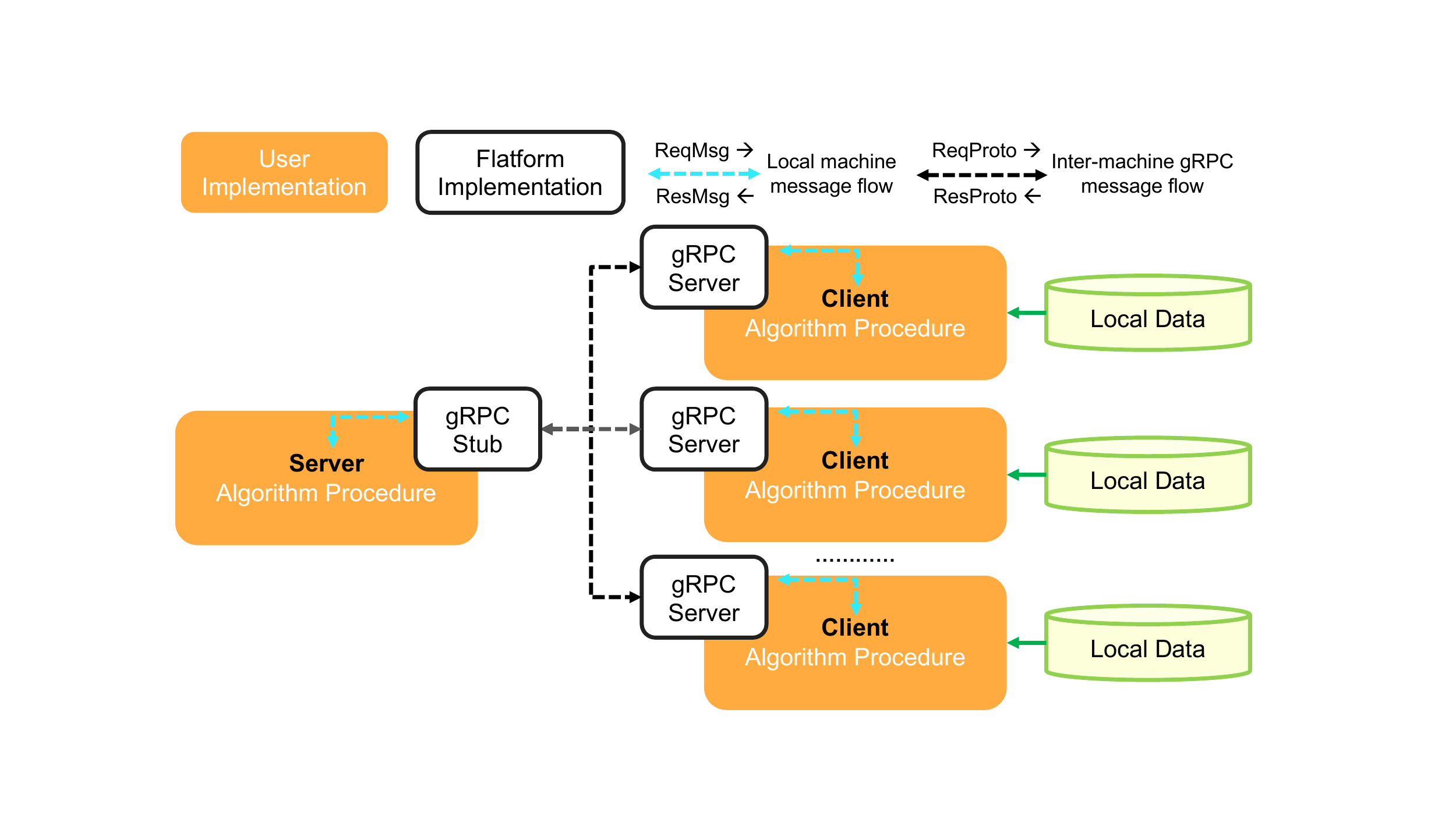}
\caption{An high-level illustration of the Fedlearn-Algo design. We provide a uniform gPRC communication module, including request message data structure, response message data structure and machine communication function call interface. Users can use it in their algorithm implementation. We provide algorithm examples to demonstrate its use.}
\label{fig:framework}
\end{figure}

A high level description of current Fedlearn-Algo is illustrated in Figure~\ref{fig:framework}. Specifically, an algorithm implemented by Fedlearn-Algo is composed of two components, platform implementation part and user implementation part. The platform implementation part contains several common components shared by all algorithms, including machine communication module (e.g. gRPC Stub and gRPC Server) and a algorithm pipeline template. User implementation part mainly contains the algorithm specific modules. We will introduce the provided vertical FL algorithm examples in \S\ref{sect:Examples}. In this part we describe the overall design of the platform implementation part.

\textbf{Machine communication}. We design two message data structures \textit{RequestMessage} and \textit{ResponseMessage}. They are used to transfer information between server and clients in all implemented algorithms. Each message contains four variables, \textit{sender}, \textit{receiver}, \textit{body} and \textit{phase\_id}. The message body is designed to be a dictionary data structure. It supports transferring multiple information in one message. An uniform function call \textit{SendMessage} is provided as the data transfer API by which the \textit{RequestMessage} can be delivered from the \textit{sender} to the \textit{receiver}. An \textit{ResponseMessage} containing the \textit{receiver}'s response is sent back to the \textit{sender} after client finish its computation.

\textbf{Algorithm pipeline template.} A federated model training process can be generally partitioned into three stages, training initialization, training loop and training wrapping up (e.g. model saving etc.). For most FL algorithms, one iteration of the training loop contains several communication rounds. We define a \textit{phase\_id} variable in \textit{RequestMessage} and \textit{ResponseMessage} to indicate the status of the corresponding communication round. The pattern is that the computation that server (client) needs to conduct can be identified by the \textit{phase\_id} it received from the client (server). We are motivated by this pattern to design a generic training control pipeline template. For each specific algorithm's implementation, a map between \textit{phase\_id} symbols and operation function needs to be defined in the function . The use is exemplified by the released vertical FL examples kernel binary classification algorithm and random forest algorithm.

\section{Exemplar Algorithms}
\label{sect:Examples}
\subsection{Vertical federated kernel binary classification}
\begin{algorithm}\caption{Federated kernel binary classification model training algorithm.}
\label{alg:kernel}
\small
\SetKwInOut{Input}{Input}
\SetKwInOut{Output}{Output}
\SetKw{Initialization}{Initialization}
\Input{Pre-defined kernel feature mapping $\phi_1, \phi_2, ...,\phi_p$. Distributed training data $X_1, X_2,..., X_P,$ where $X_p=\{x_{i,p}\}_{i=1}^N$, $p=1,2,...,P$. Training set ground truth on active party $Y=[y_1,...,y_N]^\intercal$.}
\Initialization{Initialize model parameters $w_1^{0},w_2^{0},...,w_p^{0}$.} \\
\For{$p \in \{1,2,...,P\}$ in parallel}{
Apply $\phi_p$ to $X_p$, get $\phi_p(X_p)$.
}

\For{$t=0,1,...,t_{max}$}{
\tcc{\textbf{For clients: the selected client updates model parameter by solving a linear regression task.}}
\For{$p\in \{1,2,...,P\}$ in parallel}{
If $t=0$, $cp\neq p$ and $\phi_p(X_p)$ is not null, send $\phi_p^\intercal(X_p)w_p^{(t)}$ directly, otherwise compute
\[
s_p^{(t-1)}=v^{(t-1)}-\phi_p^\intercal(X_p) w_p^{(t-1)}\]
\begin{equation}
\label{eqn:locallr}
w_p^{(t)} =\arg\min\limits_{w_p} \frac{1}{N}\|\phi_p^\intercal(X_p) w_p - s_p^{(t-1)} \|^2
\end{equation}
If $p$ is not active party, send $\phi_p^\intercal(X_p)w_p^{(t)}$ to master, otherwise send $\phi_p^\intercal(X_p)w_p^{(t)}-Y$ to master.
}
\tcc{\textbf{For master: compute sum of inner product}}
Master machine compute \begin{equation}
\label{eqn:sum}
v^{(t)}=\phi_1^\intercal(X_1) w_1^{(t)}+\phi_2^\intercal(X_2)w_2^{(t)}+...+\phi_P^\intercal(X_P) w_P^{(t)}-y
\end{equation}
then send it to all parties, then assign one party for parameter update by setting $cp$.
}
\Output{Model parameters $w_1,w_2,...,w_P$.}
\end{algorithm}
Kernel method is an classical machine learning algorithm. Given a sample $x\in R^d$, a kernel mapping $\psi$ transforms $x$ into a high dimension space such that in that feature space samples from different categories are more linearly separable. To alleviate the high dimension of kernel mapping, \citep{rahimi2007random} proposes to approximate the kernel mapping with random feature mappings, such that the kernel evaluation of two samples can be approximated by the inner product of the transformed sample, that is
\[
k(x_1, x_2) = \langle \psi(x_1), \psi(x_2) \rangle \approx \phi^\intercal(x_1) \phi(x_2)
\]
where $\phi(x)$ denotes the kernel approximation transformation.
In our example, we choose random Fourier feature approximation of RBF kernel \[\phi(x)=\sqrt{2\gamma}[\cos(z_1^\intercal x+b_1),\cos(z_2^\intercal x+b_2),...,\cos(z_D^\intercal x+b_D)]^\intercal,\]
where $z_1, z_2,..., z_D\in R^d$ are drawn from standard Gaussian distribution,  $b_1$, $b_2$,..., $b_D\in R$ are uniformly drawn from $[0, 2\pi]$, $\gamma$ is a scale parameter. The randomization property of kernel approximation algorithm make it applicable to protect the privacy of original feature. We leverage this property and propose a kernel vertical federated binary classification model.

Assume the overall training samples $X=\{(x_i,y_i)\}_{i=1}^N$ are distributed on $P$ parties and the $N$ training samples' ID have been aligned. The active party owns dataset $(X_1, Y)$, $Y=[y_1,...,y_N]^\intercal$ and other parties are passive parties with sample features $X_2, X_2,...,X_P$. The learning target is
\[
w_1,w_2,...,w_P=\arg\min\limits_{\{w_p\}_{p=1}^P} \frac{1}{N} \sum_{i=1}^N\|y_i-\sum\limits_{p=1}^P \phi^\intercal(x_{i,p})w_p\|^2
\]
where $w_p$ denotes the model parameter on the $p$-th party, $X_p=\{x_{i,p}\}_{i=1}^N$, $x_{i,p}$ denotes the $i$-th sample on the $p$-th party, $\phi(x_{i,p})$ is the kernel approximation mapping of $x_{i,p}$. For simplicity we assume $y_i\in \{-1,1\}$.

The algorithm used in this example is derived from~\citep{gu2020federated,gu2020privacy}. In~\citep{gu2020federated} a federated vertical doubly stochastic kernel learning algorithm is proposed.  \citep{gu2020privacy} proposes a asynchronous vertical federated linear model training algorithm. The algorithm updates local models on all parties in parallel. The shown example makes the following modifications for efficiency concern without losing data privacy protection measure. First, we adopt local kernel mapping on involved parties for data privacy. The random matrix and vector used for kernel approximation mapping can also encrypt the local data. Second, we adopt a batch algorithm rather than the stochastic algorithm used in~\citep{gu2020privacy} to improve training efficiency.

The training algorithm is summarized in Algorithm~\ref{alg:kernel}. First, each involved party transforms the original feature with its kernel approximation mapping function $\phi$. The training loop has two communication rounds. At the first round, one selected party updates its local model parameters by solving the local linear regression model learning task Eqn.~\ref{eqn:locallr}, then all parties send either $\phi_p^\intercal(X_p)w_p^{(t)}$ or $\phi_p^\intercal(X_p)w_p^{(t)}-Y$ to master, where
\[\phi_p(X_p)=[\phi_p(x_{1,p}),...,\phi_p(x_{N,p})].\]
At the second round, master machine aggregate the client updates via Eqn.\ref{eqn:sum} and chooses
the client for local parameter update at the next iteration, then sends the aggregation result to the clients.

\subsection{Vertical federated random forest}
\label{subsec:protocol}

\begin{algorithm}	
	\caption{Main pipeline of building one federated decision tree}
    \label{alg:main}
	\SetKwInOut{Input}{Input}
	\SetKwInOut{Output}{Output}
	\SetKw{Initialization}{Initialization}
	\Input{Feature space $F=\{F_p\}_{p=1}^P$, label set $\{y_i\}_{i=1}^N$.}
	\Initialization{Active party encrypts label and send the encrypted label $\{\left \langle y_i\right \rangle \}_{i=1}^N$ to all passive parties via server.} \\
	\For{$t=0,1,...,t_{max}$}{
	  \For {$p\in {1,2,...,P}$ in parallel}{
	     Client $p$ computes encrypted label quantile statistics $S_p$ by Algorithm~\ref{alg:sort}, then send  $S_p$ to server; }
	  Server collects $\{S_p\}_{p=1}^P$ and sends them to active party. \\
	
	  Active party find the best split parameter $(f_{opt}^t, v_{opt}^t)$ from $\{S_p\}_{p=1}^P$,
	  then send it to all other parties. \\
	  \For {$p\in {1,2,...,P}$ in parallel}{
	     If $f_{opt}^t\in F_p$, split the feature space into $(F_L^{(t)}, F_R^{(t)})$ by $f_{opt}^t$ and create child nodes.}
	}
	\Output{One decision tree}
\end{algorithm}

\begin{algorithm}
    \caption{Encrypted label quantile statistics on the $p$-th party}
    \label{alg:sort}
  	\SetKwInOut{Input}{Input}
	\SetKwInOut{Output}{Output}
	\SetKw{Initialization}{Initialization}
	\Input{Training set $X_p\in R^{N\times d_p}$. instance feature space $F_p$, feature dimension $d_p$, quantile number $l_p$, encrypted labels
	$\langle Y\rangle=\left\langle y_i\right\rangle_{i=1}^N$.}
    \For{$k = 0,...,d_p$}{
        Compute quantiles of the $k$-th dimension feature, $C_{k} = \left \{c_{k,1},c_{k,2},...,c_{k,l_p}\right \}$. \\
        \For{$v=1,...,l_p$}{
       Compute label statistics \[S_p(k,v) = \frac{1}{n_{kv}} \sum_{i \in \left \{ i|c_{k,v-1}< x_{i,k} \leq c_{k,v}\right \}}\left \langle y_{i} \right \rangle\]
       where $n_{kv}$ denotes the sample number whose feature value lies in $(c_{k,v-1}, c_{k,v}]$.
        }
    }
\end{algorithm}
Random forest (RF) is a popular tree structure model. Given a input sample $x\in R^d$, the prediction function of a RF is an ensemble of multiple decision trees:
\begin{equation}
R(x) = Agg(\{T_{i}(x)\}_{i=1}^n)
\end{equation}
where $R$ denotes the RF prediction function, $T_i$ denotes the $i$-th decision tree, $Agg$ denotes the aggregation strategy.

Because the decision trees can be trained in parallel, by proper parallel programming implementation the training efficiency of an RF model can be significantly improved. The overall training algorithm of one vertical federated decision tree is shown in Algorithm~\ref{alg:main}. We denote the training samples' feature, instance feature space and label set as $X\in R^{N\times d}$, $F$ and $Y=[y_1, ..., y_N]^\intercal$ respectively. Assume the feature space is distributed on $P$ parties, that is $X=\{X_p\}_{p=1}^P$, $F=\{F_p\}_{p=1}^P$, and there is only one party holding $Y$ as the active party. At the initialization
step, active party sends encrypted labels $\langle Y\rangle$ to all passive parties via server machine. After receiving $\langle Y \rangle$, each passive party calculates the encrypted label quantile statistics
$S_p$ via Algorithm~\ref{alg:sort}. We use $l_p$ to denote the pre-defined quantile number and $d_p$ to denote the feature dimension on $p$-th party. Therefore we have $S_p\in R^{d_p\times l_p}$ where the entry $S_p(i,j)$ denotes the average value of $\langle Y\rangle$ on the $i$-th dimension feature and $j$-th quantile. Active party receives $\{S_p\}_{p=1}^P$ from master, then evaluate which feature dimension and quantile should be used for tree split, based on proper criterion like maximum information gain. The decided feature and quantile $(f_{opt}, v_{opt})$ is sent to the corresponding party for tree split.

At the initialization step, we adopt homomorphic encryption to encrypt the labels $Y$. A good property of homomorphic encryption is that it allows for computations such as addition or multiplication on the encrypted data. Therefore we compute the label quantile statistics $\{S_p\}_{p=1}^P$ on $\langle Y\rangle$ then active party can decrypt  $\{S_p\}_{p=1}^P$ and compute the feature split based on the decrypted quantile label statistics.

\section{Conclusion and Future Work}
In this paper, we introduce Fedlearn-Algo, an open-source privacy-preserving machine learning algorithm platform. As the first part of release, we open-source two novel vertical FL models, kernel binary classification model and vertical FL random forest model. The platform is naturally compatible to existing machine learning tools (e.g. TensorFlow, PyTorch, Sklearn, etc.), by which researchers and contributors can implement their own algorithms. We believe the agnostic data format transfer interface and the algorithm template are flexible and easy-to-use.

\begin{figure}
\centering
\includegraphics[width=1\textwidth]{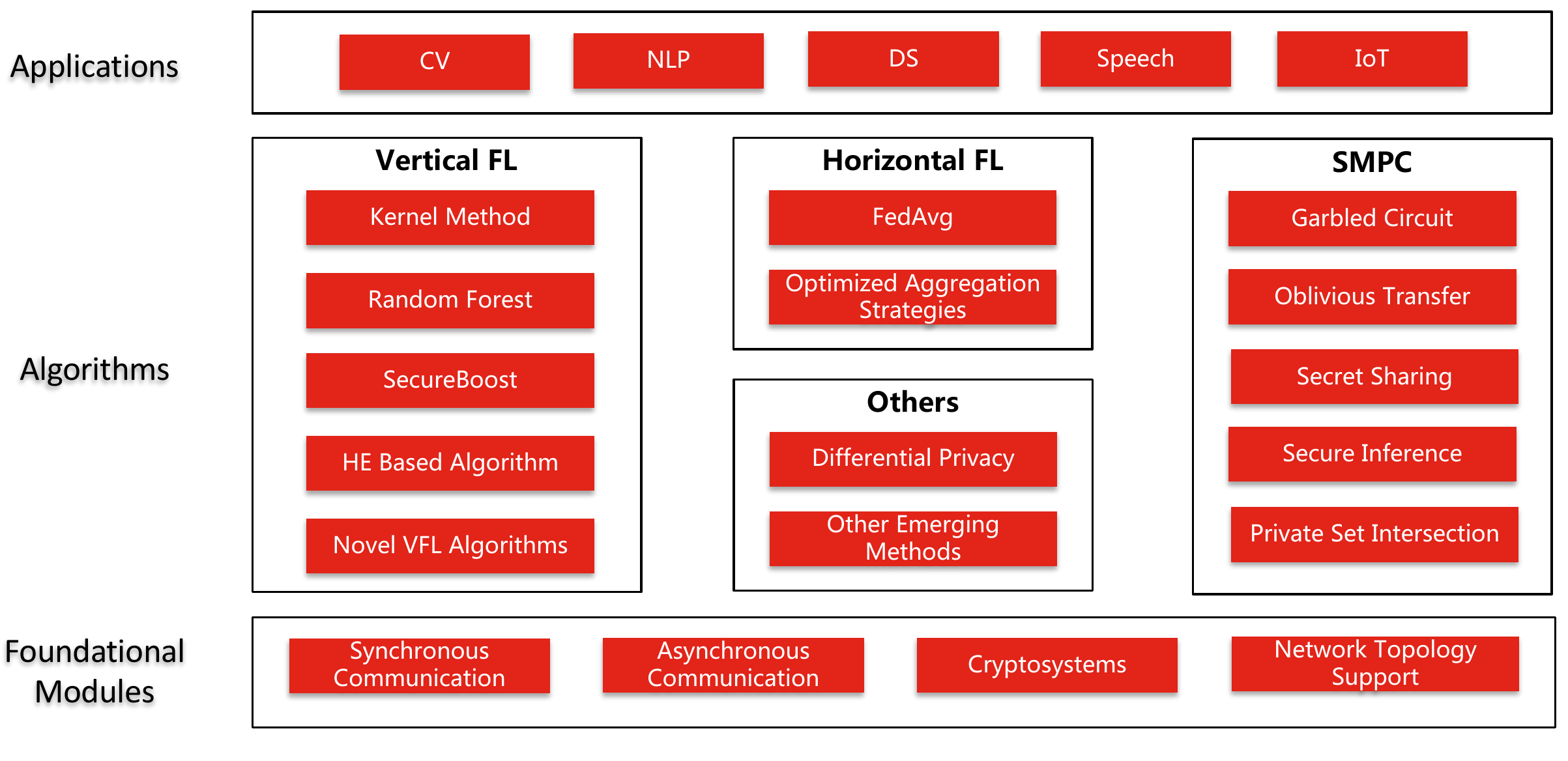}
\caption{An overall working plan of Fedlearn-Algo.}
\label{fig:plan}
\end{figure}

In the future, we will continue adding more functionality modules to Fedlearn-Algo. Our overall plan is shown in Figure~\ref{fig:plan}. Specifically, our future efforts include but are not limited to the following aspects.

\begin{itemize}
\item \textbf{Adding more functional module support.} We are working on adding asynchronous machine communication support and decentralized network topology support. We also plan to build an data encryption module, providing standard data cryptography algorithm implementations for user.

\item \textbf{Releasing our novel algorithm research results.} This platform is used to demonstrate our current and future algorithm research and development results, with emphasis on vertical FL algorithms. We will release those algorithm implementations on this platform in the future.

\item \textbf{Providing standard algorithm implementations.} Apart from the novel algorithm release, we will also provide standard privacy-preserving algorithm implementations, such horizontal FL algorithms, SMPC protocols, differential privacy methods and emerging methods such as distillation based method (e.g.~\citep{wang2019private}) and graph federated learning (e.g.~\citep{meng2021cross}).

\item \textbf{Applying privacy-preserving ML to specific use cases.} Leveraging privacy preserving ML for cross-device data use is being observed in more and more domains~\citep{pokhrel2020federated,khan2021federated}. Our team is currently exploiting the use in DS, CV and NLP. We will also consider other applications such as Speech and IoT, etc.

\end{itemize}

\small
\bibliography{reference}
\bibliographystyle{icml2017}
\end{document}